\documentclass[letterpaper,10pt,conference]{ieeeconf}

\IEEEoverridecommandlockouts
\usepackage{cite}
\usepackage{amsmath}
\usepackage{amssymb}
\usepackage{graphicx}
\usepackage{booktabs}
\usepackage{float}
\usepackage{hyperref}

\newcommand{\method}{DUAC }

\title{\LARGE \bf
Adaptive Action Horizon Selection via Flow-Field Uncertainty
}

\author{Idan Ben-Ami$^{1}$, Arnon Netzer$^{1}$, Sarah Keren$^{2}$, Idit Diamant$^{1}$\\
$^{1}$Arm AI Research (AAIR), Arm Ltd \qquad
$^{2}$Technion Israel Institute of Technology
\thanks{Idan Ben-Ami$^{1}$, Arnon Netzer$^{1}$, Sarah Keren$^{2}$, Idit Diamant$^{1}$;
$^{1}$Arm AI Research (AAIR), Arm Ltd; $^{2}$Technion Israel Institute of Technology, idit.diamant@arm.com}}

\begin{document}

\maketitle
\thispagestyle{empty}
\pagestyle{empty}

\begin{abstract}
Reliable uncertainty estimates are important for adaptive execution in
vision-language-action (VLA) models. Existing approaches such as
Adaptive Action Chunking (AAC) estimate action entropy from multiple sampled action chunks, introducing additional inference cost and capturing both uncertainty within an action trajectory and diversity across alternative strategies. We introduce Divergence-based Uncertainty-aware Adaptive Action Chunking (DUAC), a post-hoc method that extracts trajectory-conditioned predictive uncertainty from the vector field of a pretrained VLA. Building on a closed-form flow-posterior covariance formulation, DUAC derives action-level uncertainty scores from a single generated chunk using Jacobian--vector products, without retraining or multiple endpoint samples. DUAC strongly correlates with AAC, while capturing distinct information: AAC measures global dispersion across sampled chunks, whereas DUAC measures local uncertainty around the trajectory selected for execution. Consequently, DUAC is less sensitive to diversity among distinct but valid action strategies. 
Across LIBERO, LIBERO-Pro, LIBERO-Plus, and RoboCasa, DUAC-based adaptive horizon selection performs favorably among the evaluated methods, combining strong task success with low uncertainty-estimation latency.
These results establish
the flow vector field as an efficient, trajectory-conditioned source of
predictive uncertainty for robot decision making.
\end{abstract}

\section{Introduction}

Vision-language-action (VLA) models have emerged as a general approach to robotic manipulation, mapping visual observations, language instructions, and robot state into executable actions~\cite{brohan2022rt1,zitkovich2023rt2,kim2024openvla,octo2024,kim2025openvlaoft,bu2025univla}. Many recent VLAs generate continuous action chunks with diffusion or flow-matching action heads~\cite{octo2024,li2024cogact,liu2025rdt,hou2025ditpolicy,black2024pi0,black2025pi05,bjorck2025groot,shukor2025smolvla}. Starting from noise, these policies repeatedly evaluate a learned vector field to transform a latent variable into a coherent action trajectory. Despite the inherently iterative nature of this computation, the internal flow trajectory is usually treated only as a mechanism for producing the final action, rather than as a source of information about the reliability of that action.

\begin{figure}[!t]
  \centering
  \includegraphics[width=0.9\columnwidth]{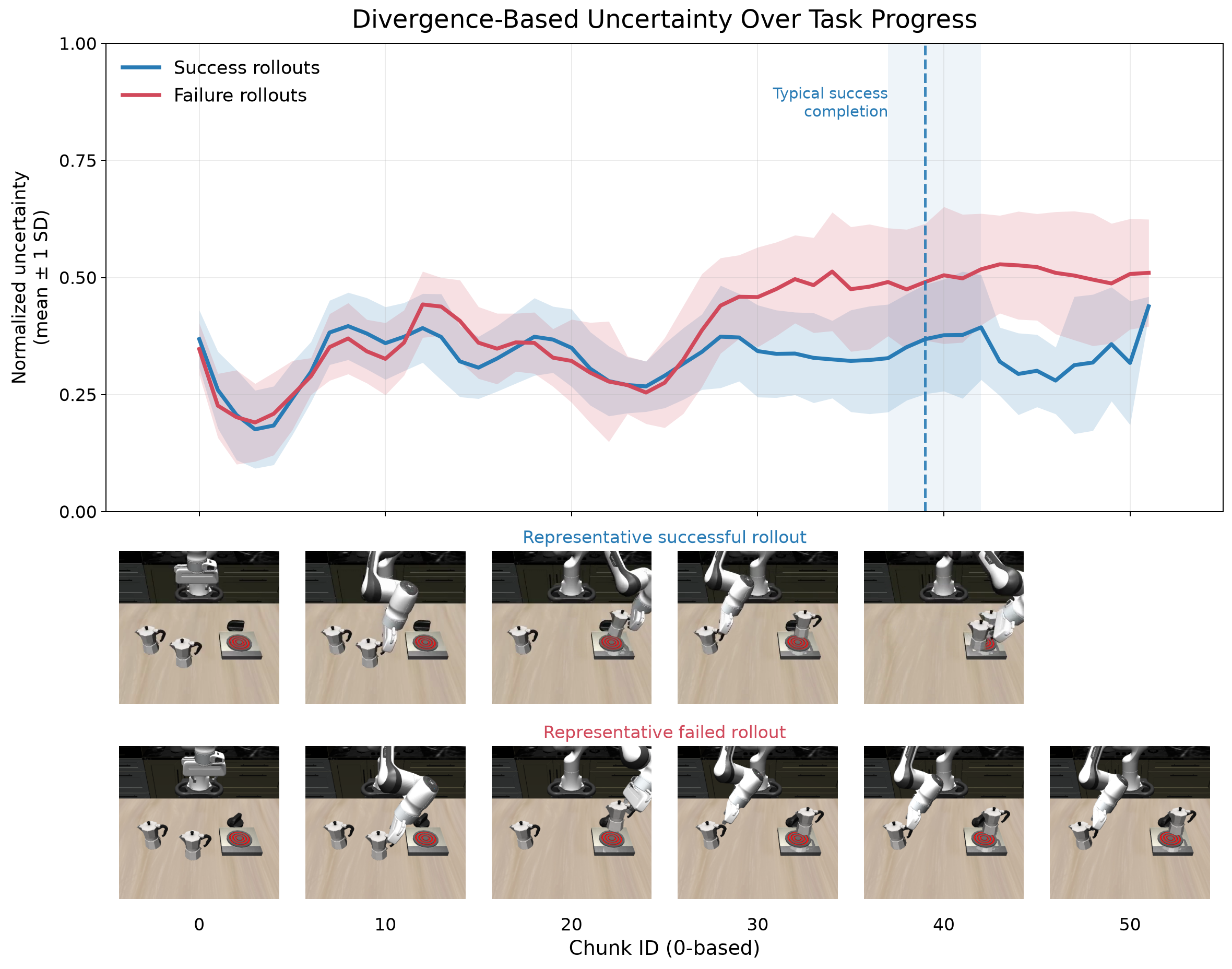}
  \caption{Uncertainty over task progress for successful and failed LIBERO-10 rollouts. Curves show mean $\pm1$ standard deviation. 
  Frame rows show representative outcomes.}
  \label{fig:task8_progress_v3}
\end{figure}

Reliable estimates of predictive action uncertainty are particularly useful when action chunks are executed in closed loop~\cite{chen2026reconvla,liang2026aac,wang2026vlaknows}. Longer execution horizons preserve temporal consistency and reduce the frequency of policy inference, but limit the policy's responsiveness to new observations. Shorter horizons permit frequent replanning, but can produce discontinuous or overly reactive behavior. Adaptive Action Chunking (AAC)~\cite{liang2026aac} addresses this trade-off by sampling multiple candidate action chunks, estimating the entropy of the resulting action distribution at each predicted step, and shortening the execution horizon when this entropy increases. While effective, the quality and stability of its estimate depend on the number of sampled chunks, and obtaining multiple complete action trajectories introduces additional inference-time computation and memory.

These limitations motivate using the local geometry of a flow-matching policy to estimate predictive action uncertainty without sampling multiple terminal action chunks. Xing et al.~\cite{xing2026divergence} derive a closed-form expression that relates the conditional posterior covariance of a flow-matching model to the divergence of its learned vector field. However, the utility of this formulation for VLA action generation remains unexplored, including whether it can provide uncertainty for individual actions within a predicted chunk and support adaptive action horizon execution.

We introduce \textbf{D}ivergence-based \textbf{U}ncertainty-aware Adaptive \textbf{A}ction \textbf{C}hunking, \textbf{\method}, a post-hoc method that extracts predictive uncertainty from the vector field of a pretrained flow-matching VLA. We adapt the flow-posterior covariance formulation to  VLA policies and derive an action-wise uncertainty signal for generated action chunks. At an intermediate state of the flow process, \method estimates the dispersion of terminal actions associated with that local flow state. When nearby action trajectories converge toward similar terminal actions, the prediction is more certain. When they remain dispersed, the terminal action is less determined. Unlike entropy-based uncertainty in AAC~\cite{liang2026aac}, which summarizes variability across independently sampled action chunks, our score is conditioned on the intermediate state of the specific trajectory being generated. It therefore provides a local, trajectory-specific uncertainty estimate for the action chunk selected for execution. Figure~\ref{fig:task8_progress_v3} illustrates this signal, showing higher uncertainty during later stages of unsuccessful rollouts.

We compare DUAC with the sampling-based action entropy used by AAC~\cite{liang2026aac} and observe strong correlations in both their per-action uncertainty profiles and selected execution horizons. Nevertheless, the two methods capture different information: AAC measures global dispersion across sampled action chunks, whereas DUAC measures local uncertainty around the particular trajectory selected for execution. Consequently, DUAC is less sensitive to variability among distinct but valid action strategies and avoids treating such diversity as uncertainty in the selected trajectory. Across LIBERO, LIBERO-Plus, LIBERO-Pro, and RoboCasa, DUAC improves performance while substantially reducing uncertainty-estimation latency relative to entropy-based AAC.
Our contributions are summarized as follows:
\begin{itemize}
    \item We introduce Divergence-based Uncertainty-aware Adaptive Action Chunking (DUAC), which adapts the closed-form flow-posterior covariance to VLA action generation and derives per-action uncertainty scores from one generated action chunk using Jacobian--vector products.

    \item We provide a sampling-efficient alternative to entropy-based AAC that requires one generated action chunk and achieves substantially lower uncertainty-estimation latency in our implementation.

    \item Since DUAC captures local uncertainty around the selected action trajectory, we show that it is less sensitive than AAC's entropy to variability among distinct but valid action strategies, reducing unnecessary replanning.

    \item We apply DUAC to adaptive horizon selection, achieving improved performance over the evaluated methods.
\end{itemize}

\begin{figure*}[!t]
  \centering
  \includegraphics[height=150pt, width=\linewidth]{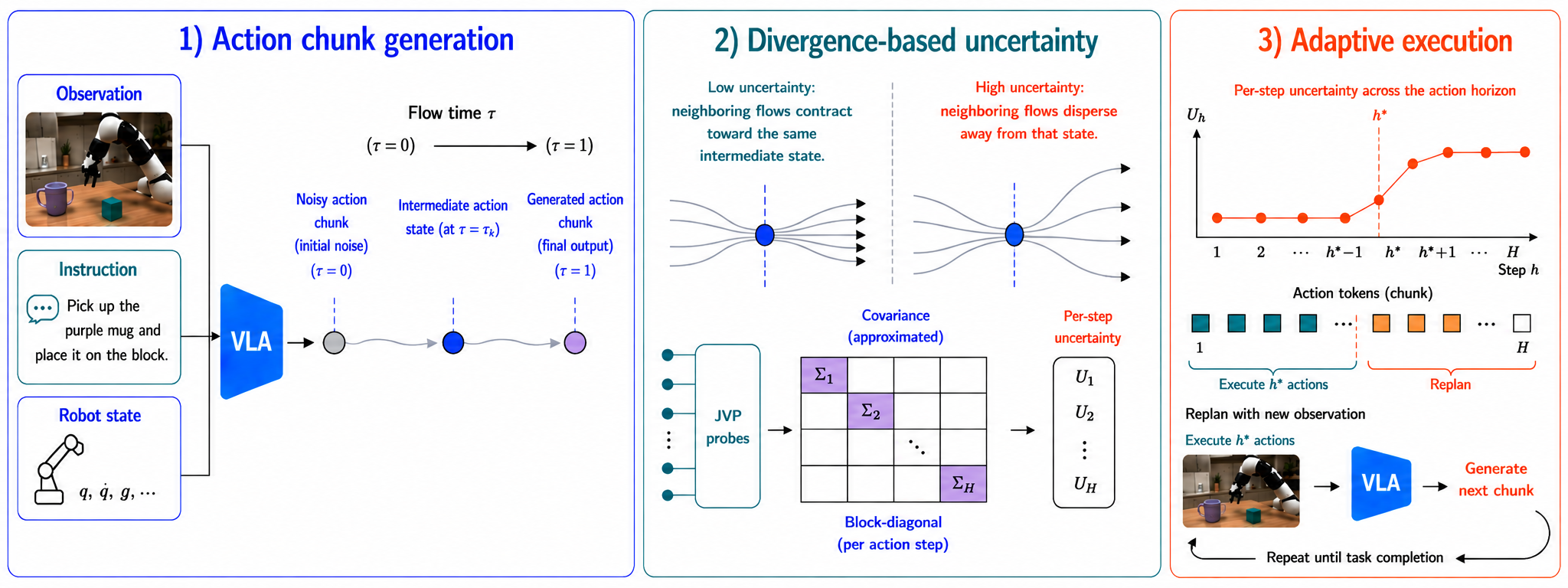}
  \caption{Method overview of divergence-based uncertainty estimation for flow-based action generation and its use in adaptive execution-horizon selection.}
  \label{fig:method_overview}
\end{figure*}

\section{Method}

Our method extracts predictive uncertainty from an action chunk generated by a flow-matching policy. We build on the recent work of Xing et al.~\cite{xing2026divergence}, who show that the predictive uncertainty of a flow-matching model is encoded in the local geometry of its learned velocity field. We adapt this formulation to derive per-action uncertainty scores, yielding an uncertainty profile over the prediction horizon. We use this profile to determine how many actions to execute before replanning. Figure~\ref{fig:method_overview} summarizes the uncertainty estimation and its use for adaptive action execution. The method requires no additional training or multiple action samples.

\subsection{Action-Chunk Divergence-Based Uncertainty}

At robot decision \(n\), the policy predicts an action chunk
\begin{equation}
a_n=[a_{n,1},\ldots,a_{n,H}]
\in\mathbb{R}^{H\times D_a},
\end{equation}
where \(H\) is the prediction horizon and \(D_a\) is the action dimension.
Under the standard linear flow-matching path~\cite{lipman2023flowmatching},
\(a_{n,\tau}=(1-\tau)\epsilon_n+\tau a_n\), where
\(\epsilon_n\sim\mathcal{N}(0,I)\) and \(\tau\in[0,1]\). Thus, \(\tau=0\)
corresponds to noise and \(\tau=1\) to the generated action chunk; additional
background is provided in Appendix~\ref{supp:preliminaries}.
The vector field \(v_\theta(a_{n,\tau},\tau,c_n)\) is conditioned on the
observation, instruction, and robot state \(c_n\). After vectorizing the chunk
in \(\mathbb{R}^{HD_a}\), we apply the closed-form posterior-covariance identity
of Xing et al.~\cite{xing2026divergence}:
\begin{equation}
\operatorname{Cov}
\left(a_n\mid a_{n,\tau},c_n\right)
=
\frac{(1-\tau)^2}{\tau}
\left[
I_{HD_a}+(1-\tau)J_{v_\theta}
\right],
\label{eq:action_chunk_covariance}
\end{equation}
where \(I_{HD_a}\) is the identity matrix, and 
\(J_{v_\theta}=\partial v_\theta/\partial a_{n,\tau}\). For the learned
vector field, we treat this expression as a model-implied covariance estimate.
Taking the trace yields the following chunk-level uncertainty:
\begin{equation}
U_{\mathrm{chunk}}(\tau)
=
\frac{(1-\tau)^2}{\tau}
\left[
HD_a+(1-\tau)\operatorname{div}v_\theta
\right].
\label{eq:chunk_uncertainty}
\end{equation}
where \(\operatorname{div}v_\theta=\operatorname{Tr}(J_{v_\theta})\).

\subsection{Per-Action Divergence-Based Uncertainty}

Adaptive horizon selection requires an uncertainty score for each future action rather than a single score for the complete action chunk. Each predicted action \(a_{n,i}\in\mathbb{R}^{D_a}\) contains \(D_a\) action dimensions. Accordingly, the conditional covariance
\(\operatorname{Cov}(a_n\mid a_{n,\tau},c_n)\) can be viewed as consisting of \(H\times H\) blocks, where each block
is
$
\Sigma_{ij}
=
\operatorname{Cov}
\left(
a_{n,i},a_{n,j}
\mid a_{n,\tau},c_n
\right).
$
The diagonal block associated with action \(i\), $\Sigma_{ii}$, 
contains the conditional variances and within-action covariances of the \(D_a\) components of \(a_{n,i}\).
To connect these covariance blocks to the vector field, we similarly partition its Jacobian according to action positions:
\begin{equation}
J_{ij}
=
\frac{\partial v_\theta^{(i)}}
{\partial a_{n,\tau}^{(j)}}
\in\mathbb{R}^{D_a\times D_a},
\label{eq:block_jacobian}
\end{equation}
where \(v_\theta^{(i)}\) is the vector-field output associated with the \(i\)-th future action, and \(a_{n,\tau}^{(j)}\) is the intermediate action at position \(j\). The block \(J_{ij}\) measures how the vector-field output for action \(i\) changes when the \(j\)-th intermediate action is perturbed.

Applying the closed-form covariance identity to the \(i\)-th diagonal block gives
\begin{equation}
\Sigma_{ii}(\tau)
=
\frac{(1-\tau)^2}{\tau}
\left[
I_{D_a}
+
(1-\tau)
\frac{\partial v_\theta^{(i)}}
{\partial a_{n,\tau}^{(i)}}
\right].
\label{eq:step_covariance}
\end{equation}
We obtain a scalar uncertainty score for action \(i\) by taking the trace of this block:
\begin{equation}
\begin{aligned}
U_i(\tau)
&=
\operatorname{Tr}\left[\Sigma_{ii}(\tau)\right] \\
&=
\frac{(1-\tau)^2}{\tau}
\left[
D_a
+
(1-\tau)
\operatorname{Tr}
\left(
\frac{\partial v_\theta^{(i)}}
{\partial a_{n,\tau}^{(i)}}
\right)
\right].
\end{aligned}
\label{eq:step_uncertainty}
\end{equation}
Thus, \(U_i(\tau)\) sums the conditional variances of the \(D_a\) components of action \(a_{n,i}\). Given the intermediate chunk \(a_{n,\tau}\) and context \(c_n\), the posterior \(p(a_n\mid a_{n,\tau},c_n)\) represents the possible terminal action chunks that remain compatible with the current flow state. 
The block \(\Sigma_{ii}\) captures the variability of the \(i\)-th action across these possible chunks, providing the per-action predictive uncertainty used to determine how many actions can be executed before replanning.

The trace of the full action-chunk covariance decomposes into the traces of its diagonal action blocks. Consequently, the per-action uncertainty scores recover the chunk-level uncertainty:
\begin{equation}
U_{\mathrm{chunk}}(\tau)
=\operatorname{Tr}\!\left[\operatorname{Cov}
\left(a_n\mid a_{n,\tau},c_n\right)\right]
=\sum_{i=1}^{H}U_i(\tau).
\label{eq:step_decomposition}
\end{equation}
The resulting sequence $[U_1(\tau),\ldots,U_H(\tau)]$
provides an uncertainty profile over the predicted horizon.
Because horizon selection requires uncertainty estimates for the actions that are close to execution, we use \(\tau=0.9\) in all experiments.

We estimate each divergence term using Hutchinson's trace estimator~\cite{hutchinson1990stochastic} with $K$ Rademacher probes. Jacobian-vector products are computed through automatic differentiation, avoiding explicit construction of the full action Jacobian. The method can therefore be applied post hoc to a pretrained policy without retraining.

\subsection{Adaptive Execution Horizon}

Following AAC~\cite{liang2026aac}, for each prefix length
\(h\in\{1,\ldots,H\}\), we compute the mean uncertainty
\(\overline{U}_h=\frac{1}{h}\sum_{i=1}^{h}U_i\).
Then we select the point at which the prefix uncertainty increases most sharply:
\begin{equation}
h^*
=
\max\left(
\operatorname*{arg\,max}_{1\leq h<H}
\left(\overline{U}_{h+1}-\overline{U}_h\right),
\,\xi
\right),
\end{equation}
where $\xi$ is the AAC lower bound that enforces a minimum execution horizon. The policy executes the first $h^*$ actions and then replans from a new observation.

\subsection{Relation to sampling-based uncertainty}
DUAC and AAC measure related but distinct uncertainty quantities. AAC estimates marginal action entropy from the sample covariance of multiple generated action
chunks, whereas DUAC estimates predictive covariance conditioned on an
intermediate state of one generated trajectory. Under the linear
Gaussian flow interpolation, the law of total covariance decomposes
the marginal action covariance into the expected trajectory-conditioned
covariance and the variability of the conditional means across
trajectories. DUAC captures the former component and is therefore less
sensitive to diversity among distinct but valid action strategies. 
We provide the complete derivation in
Appendix~\ref{supp:local-global-uncertainty}.

\section{Experiments}
Our experiments address the following questions: (i) Can \method achieve competitive or improved task success while reducing the latency and computational cost of adaptive action execution? (ii) What information does \method which is local, trajectory-specific divergence capture compared with the global, distribution-level uncertainty measured by AAC~\cite{liang2026aac}, particularly under distribution shift?

\subsection{Experimental Setup}
We evaluate publicly available $\pi_{0.5}$ policies~\cite{black2025pi05} on
standard LIBERO, LIBERO-Pro, LIBERO-Plus, and RoboCasa Kitchen-24. We compare
\method with fixed-horizon execution, AutoHorizon~\cite{wang2026vlaknows}, and
entropy-based AAC~\cite{liang2026aac}. All uncertainty estimates are computed
from pretrained policies at inference time, without retraining. Appendix~\ref{supp:experimental_setup} provides complete
benchmark and evaluation details.

\begin{table}[t]
  \centering
  \footnotesize
  \setlength{\tabcolsep}{6pt}
  \resizebox{\columnwidth}{!}{%
  \begin{tabular}{lccccc}
    \toprule
    Method & $x0.1$ & $x0.2$ & $x0.3$ & $x0.4$ & Overall \\
    \midrule
    Fixed-10 & 89.8\% & 54.7\% & 34.0\% & 10.2\% & 47.2\% \\
    AutoHorizon & 89.4\% & 54.4\% & 34.9\% & 10.4\% & 47.3\% \\
    AAC & 90.1\% & \textbf{57.8\%} & 36.3\% & 10.6\% & 48.7\% \\
    \method (ours) & \textbf{90.3\%} & \textbf{57.8\%} & \textbf{37.1\%} & \textbf{11.7\%} & \textbf{49.2\%} \\
    \bottomrule
  \end{tabular}%
  }
  \caption{LIBERO-Pro results: success rate (\%) under increasing x-position perturbations.}
  \label{tab:libero_pro_control_v3}
\end{table}

\subsection{Performance comparison}

We investigate whether divergence-based uncertainty provides a signal for adaptive action execution. Following AAC~\cite{liang2026aac}, the execution horizon is selected from the per-step uncertainty profile. The action prefix is executed before the policy replans from a new observation. Because this decision applies to the specific actions about to be executed, we evaluate the profile at a late flow time, when the trajectory is largely formed but retains nonzero residual uncertainty. 
We use \(\tau=0.9\), which shows the strongest agreement with entropy-based
uncertainty in supplementary Figure~\ref{fig:entropy_divergence_v3}.

\textbf{Standard LIBERO.}
We evaluate LIBERO, where fixed-horizon performance is already near saturation. Supplementary Table~\ref{tab:libero_control_v3}   shows that \method achieves similar (slightly higher) average success rate and the best results on LIBERO-10.

\textbf{Performance under distribution shift.}
Table~\ref{tab:libero_pro_control_v3} shows that \method achieves the highest
overall success under LIBERO-Pro perturbations. Its advantage is small under
mild shifts but becomes clearer under moderate and strong perturbations, where
both uncertainty-based methods outperform fixed-horizon execution. This
suggests that adaptive replanning becomes most useful once the shift
meaningfully affects action prediction. Across LIBERO-Plus categories
(Table~\ref{tab:libero_plus_control_v3}), \method achieves the highest average,
with its clearest gain under camera perturbations.

\begin{table}[t]
  \centering
  \footnotesize
  \setlength{\tabcolsep}{3pt}
  \begin{tabular}{lccc|ccc}
    \toprule
    & \multicolumn{3}{c|}{Control Performance} 
    & \multicolumn{3}{c}{Action chunk generation} \\
    \cmidrule(lr){2-4} \cmidrule(lr){5-7}
    Method & Success & Sim. steps & Decisions 
           &  Latency (ms) & Rel. cost \\
    \midrule
    Fixed-10 
        & 47.20\% & 230.20 & 22.24 
         & 68.26 & $1.0\times$ \\
    AAC 
        & 48.70\% & 229.91 & 30.60 
         & 582.56 & $8.5\times$ \\
    \method (ours)
        & \textbf{49.20\%} & \textbf{228.86} & 27.75 
        & \textbf{204.09} & \textbf{$3.0\times$} \\
    \bottomrule

  \end{tabular}

  \caption{
    Performance and computational efficiency comparison on LIBERO-Pro. 
  }
  \label{tab:libero_pro_performance_efficiency}
\end{table}

\textbf{RoboCasa Kitchen-24.}
RoboCasa exposes task-dependent horizon preferences (Table~\ref{tab:robocasa_control_v3}): Fixed-25 performs well on
precision-sensitive relocation and button tasks, whereas Fixed-50
improves rotation but lowers overall success. \method accommodates
these different control requirements, achieving the highest overall
and rotation success rates while remaining competitive across the
other task categories. Moreover, \method requires
$\sim{15}\%$ fewer policy decisions per episode than Fixed-25 while achieving
a higher overall success rate.

\begin{table}[t]
  \centering
  \footnotesize
  \setlength{\tabcolsep}{1.3pt}
  \resizebox{0.94\columnwidth}{!}{%
  \begin{tabular}{lccccccc}
    \toprule
    Method & Reloc. & Contact & Rot. & Button
    & Overall & Mean H & Dec./ep. \\
    \midrule
    Fixed-25
    & \textbf{58.80\%} & 65.33\% & 26.80\% & \underline{43.00\%}
    & \underline{51.79\%} & 25.00 & 18.98 \\
    Fixed-50
    & 54.90\% & 65.83\% & \underline{30.80\%} & 32.33\%
    & 49.79\% & 50.00 & 9.67 \\
    AutoHorizon
    & 51.60\% & \textbf{68.83\%} & 27.80\% & 37.67\%
    & 49.21\% & 34.96 & 13.84 \\
    Entropy AAC
    & 51.20\% & 59.83\% & 25.80\% & \textbf{43.67\%}
    & 47.12\% & 26.80 & 18.43 \\
    DUAC (ours)
    & \underline{57.20\%} & \underline{66.67\%} & \textbf{31.00\%} & \underline{43.00\%}
    & \textbf{52.33\%} & 29.58 & 16.17 \\
    \bottomrule
  \end{tabular}
  }
  \caption{RoboCasa Kitchen-24 results: success rate (\%), mean execution horizon (Mean H), and mean policy decisions per episode (Dec./ep.).}
  \label{tab:robocasa_control_v3}
\end{table}

\textbf{Estimator efficiency. } AAC~\cite{liang2026aac} approach requires multiple candidate action chunks, whereas \method estimates uncertainty along a single generated trajectory. Table~\ref{tab:libero_pro_performance_efficiency} summarizes both computational and control-level efficiency on LIBERO-Pro. In our JAX/OpenPI implementation on an NVIDIA A100 with 10 denoising steps, AAC uses 20 candidate action chunks, while \method uses 40 Hutchinson probes, reducing uncertainty-estimation latency by ${\sim 2.9}\times$. At the control level, \method achieves the highest success rate and lowest mean simulator steps with fewer policy decisions than AAC approach.

\begin{figure}[t]
  \centering
  \includegraphics[width=\columnwidth]{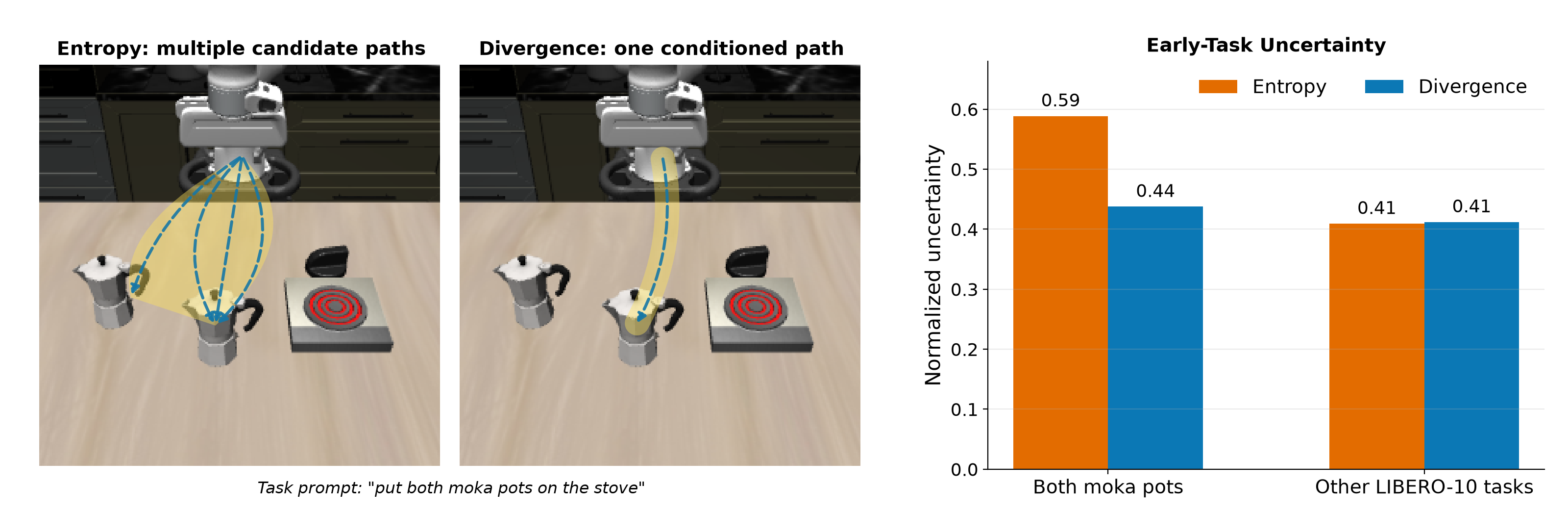}
  \caption{Multiple valid strategies increase marginal entropy more strongly than trajectory-conditioned divergence. The illustrations contrast alternative sampled trajectories with uncertainty around one selected trajectory; the right panel reports normalized uncertainty over the first three chunks. 
  }
  \label{fig:moka_order_ambiguity_v3}
\end{figure}

\subsection{Uncertainty information captured by \method}
\textbf{Uncertainty under multiple valid strategies.}
AAC estimates marginal entropy across sampled chunks and therefore responds
to variation among valid action strategies, whereas DUAC estimates uncertainty
around the trajectory selected by the flow. In the ``put both moka pots on the
stove'' task, samples may choose different pots or approaches, increasing
entropy even when each trajectory is internally concentrated. As shown in
Figure~\ref{fig:moka_order_ambiguity_v3}, AAC entropy increases substantially
relative to other LIBERO-10 tasks, whereas DUAC changes only modestly. This
suggests that AAC captures both local dispersion and strategy diversity, while
DUAC more selectively characterizes the chosen trajectory.

\begin{figure}[!t]
  \centering
  \includegraphics[width=0.74\columnwidth]{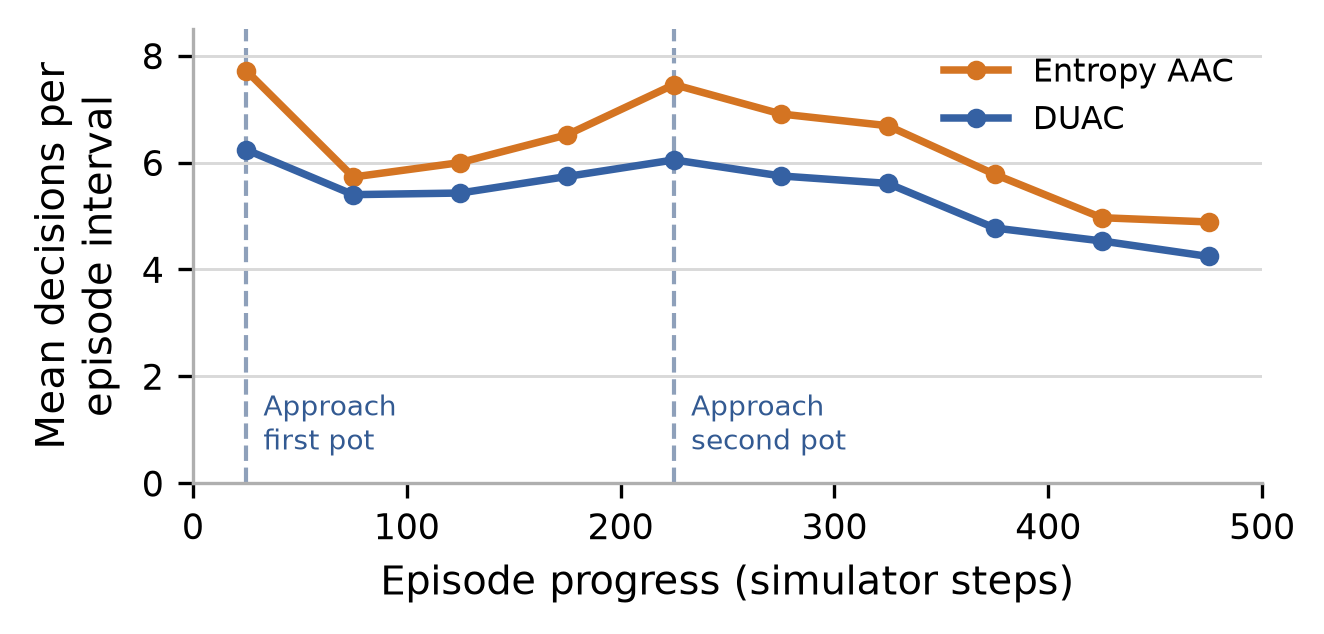}
  \caption{Mean policy decisions over task progress.}
  \label{fig:task8_decisions_v3}
\end{figure}

\textbf{Effect on adaptive execution.}
On the two-moka-pot task, DUAC replans less frequently while matching AAC's
\(75\%\) success rate. It increases the mean horizon from \(7.53\) to \(8.83\)
actions and reduces policy decisions by \(14.7\%\), suggesting that
trajectory-conditioned uncertainty avoids unnecessary replanning caused by
valid strategy diversity. Figure~\ref{fig:task8_decisions_v3} shows this trend
over task progress; Supplementary Table~\ref{tab:task8_decisions_v3} provides the
aggregate results.

\section{Discussion and Conclusion}

Our results show that the vector field of a pretrained flow-matching VLA provides a useful trajectory-conditioned uncertainty signal. Unlike AAC method, which measures dispersion across multiple sampled action chunks, DUAC estimates the uncertainty remaining around the particular trajectory selected for execution. The two measures exhibit strong agreement at late flow times, but differ when multiple valid strategies exist: marginal entropy increases with diversity across strategies, whereas DUAC more selectively characterizes the chosen trajectory. When used for adaptive horizon selection, DUAC achieves the highest observed aggregate success rates in our evaluation and reduces uncertainty-estimation latency relative to Entropy AAC.

\bibliographystyle{IEEEtran}
\bibliography{references}

\newpage
\appendix
\subsection{Experimental Setup}
\label{supp:experimental_setup}

\textbf{Benchmarks.}
We evaluate on four simulation settings. Standard LIBERO~\cite{liu2023libero}
contains the Spatial, Object, Goal, and LIBERO-10 suites and serves as our
in-distribution benchmark. LIBERO-Pro~\cite{zhou2025liberopro} introduces
controlled position perturbations of increasing severity. LIBERO-Plus
~\cite{fei2026liberoplus} covers camera viewpoint, robot initialization,
sensor noise, language, lighting, background, and object-layout perturbations.
RoboCasa Kitchen-24~\cite{nasiriany2024robocasa} contains 24 manipulation tasks
in diverse simulated kitchens and uses substantially longer action horizons
than LIBERO.

\textbf{Evaluation Protocol.}
We evaluate publicly available $\pi_{0.5}$ policies~\cite{black2025pi05}
without modifying their architecture or training. We use Physical Intelligence's
official OpenPI \texttt{pi05\_libero} checkpoint~\cite{physicalintelligence2026pi05libero}
for all LIBERO evaluations and the public RoboCasa checkpoint released by
LeRobot~\cite{lerobot2026pi05robocasa} for Kitchen-24. We conduct 50 rollouts
per task on LIBERO and 100 rollouts per task on RoboCasa. We compare \method
with three baselines: fixed-horizon execution, AutoHorizon
~\cite{wang2026vlaknows}, which adapts the horizon using action similarity, and
AAC~\cite{liang2026aac}, which uses the entropy of 20 independently sampled
action chunks. Unless stated otherwise, \method estimates the vector-field
Jacobian trace using 40 Hutchinson probes. All uncertainty estimates are
computed at inference time without retraining or auxiliary confidence modules.

\subsection{Related Work}

\textbf{Flow-Based Action Generation in VLAs.}
Recent VLAs ~\cite{black2024pi0, black2025pi05, bjorck2025groot} use conditional generative action heads to jointly synthesize continuous action chunks rather than predicting each action independently. $\pi_0$ \cite{black2024pi0} and $\pi_{0.5}$ ~\cite{black2025pi05} couples a pretrained vision-language backbone with a flow-matching action expert that transforms noise into a multi-step motor trajectory conditioned on visual observations, language instructions, and robot state. 
Similarly, GR00T ~\cite{bjorck2025groot} uses a diffusion-transformer action module, conditioned on a vision-language representation, to generate chunked controls for heterogeneous robot embodiments. 
By modeling the future actions jointly, these generative heads capture cross-timestep dependencies and multimodal action distributions, yielding temporally coherent trajectories suitable for dexterous control.

\textbf{Uncertainty-Guided VLA.}
Recent work estimates VLA uncertainty to anticipate failures, detect distribution shift, guide data acquisition, and support runtime control ~\cite{romer2026uqvla, chen2026reconvla, huang2026vlaconf, zollo2025confidence, liang2026aac, wang2026vlaknows}. R{\"o}mer et al.~\cite{romer2026uqvla} quantify epistemic uncertainty through velocity-field disagreement across an ensemble of flow-matching policies, using the resulting signal for deployment-time failure detection and uncertainty-guided active fine-tuning. ReconVLA~\cite{chen2026reconvla} applies anomaly detection in the robot state space to identify unsafe or out-of-distribution states before failure. 
Beyond failure awareness, uncertainty has also been used to determine how long a predicted action chunk should be executed before replanning. AutoHorizon~\cite{wang2026vlaknows} avoids probabilistic uncertainty estimation and instead uses action self-attention as a proxy for the policy's predictive limit. AAC~\cite{liang2026aac} estimates per-step action entropy from multiple sampled chunks and selects the execution horizon at the sharpest increase in uncertainty. 
Existing approaches provide actionable reliability signals but often require repeated sampling, policy ensembles, or auxiliary learned components. Our method instead derives uncertainty directly from the local geometry of the learned vector field along a single generated trajectory, requiring neither additional training nor multiple models or action samples.

\subsection{Background and Derivations}
\label{supp:preliminaries}

\textbf{Flow Matching.}
Flow matching learns how to transform a simple noise distribution into a data distribution. We use the standard linear interpolation~\cite{lipman2023flowmatching}
\begin{equation}
x_\tau = (1-\tau)x_0 + \tau x_1,
\end{equation}
where $x_0 \sim \mathcal{N}(0,I)$ is Gaussian noise, $x_1 \sim p_{\mathrm{data}}(\cdot \mid c)$ is a data sample, and $\tau \in [0,1]$ denotes flow time. The variable $c$ contains the conditioning information, such as the robot's observations and task instruction. Thus, $x_\tau$ moves from noise at $\tau=0$ to the data sample at $\tau=1$.

A flow-matching model learns a velocity field
$v_\theta(x_\tau,\tau,c)$ that predicts the direction from $x_0$ to $x_1$. 
At inference time, the model begins with a noise sample and follows the learned velocity field from $\tau=0$ to $\tau=1$. In an action policy, the resulting sample $x_1$ represents a predicted action or action sequence.

\textbf{Uncertainty from Flow-Field Geometry.}
Xing et al.~\cite{xing2026divergence} show that the predictive
uncertainty of a flow-matching model is encoded in the local geometry
of its learned velocity field. Under the Gaussian interpolation
assumptions, the conditional covariance of the terminal sample is
\begin{equation}
\operatorname{Cov}(x_1\mid x_\tau,c)
=
\frac{(1-\tau)^2}{\tau}
\left[
I+(1-\tau)J_{v_\theta}
\right],
\end{equation}
where \(J_{v_\theta}\) is the Jacobian of the vector field with respect
to \(x_\tau\). With a learned vector field, we treat this expression as
a model-implied estimate of the terminal-sample covariance.
Taking its trace gives the scalar predictive uncertainty:

\begin{equation}
U(x_\tau,\tau)
=
\frac{(1-\tau)^2}{\tau}
\left[
d+(1-\tau)\operatorname{div}v_\theta
\right],
\end{equation}
where $d$ is the dimensionality of the generated sample and $\operatorname{div}v_\theta=\operatorname{Tr}(J_{v_\theta})$. 
This score measures local flow contraction or expansion, with stronger contraction indicating lower predictive uncertainty.

\textbf{Divergence and Sampled-Based Uncertainty Relationship.}
\label{supp:local-global-uncertainty}
We relate the proposed local covariance to the covariance underlying AAC's continuous-action entropy estimator.
The divergence-based and sampled-based ~\cite{liang2026aac} estimators measure related but distinct notions of uncertainty. The divergence estimator measures uncertainty conditioned on a particular intermediate state $a_\tau$, while sampled-based estimator measures the total uncertainty of the endpoint action.
For the linear interpolant:
\begin{equation}
    a_\tau=\tau a+(1-\tau)\varepsilon,
    \qquad \varepsilon \sim \mathcal{N}(0,I),
\end{equation}
the likelihood of $a_\tau$ given an endpoint action $a$ is
\begin{equation}
    p(a_\tau\mid a,c)
    =
    \mathcal{N}\!\left(a_\tau;\tau a,(1-\tau)^2I\right).
\end{equation}
Hence, by Bayes' rule,
\begin{equation}
    p(a\mid a_\tau,c)
    =
    \frac{
        \mathcal{N}\!\left(a_\tau;\tau a,(1-\tau)^2I\right)p(a\mid c)
    }{
        \int
        \mathcal{N}\!\left(a'_\tau;\tau a',(1-\tau)^2I\right)p(a'\mid c)\,da'
    }.
\end{equation}
Thus, the local posterior is the global action distribution $p(a\mid c)$ reweighted according to compatibility with the current noisy state $a_\tau$.
Marginalizing over intermediate states recovers the global distribution,
\begin{equation}
    p(a\mid c)
    =
    \int p(a\mid a_\tau,c)\,p(a_\tau\mid c)\,da_\tau.
\end{equation}
Therefore, applying the law of total covariance gives
\begin{equation}
\begin{aligned}
    \Sigma_{}(c)
    =
    \mathbb{E}_{a_\tau\mid c}
    \left[
        \Sigma_{\mathrm{div}}(a_\tau,\tau,c)
    \right]
    +
    \operatorname{Cov}_{a_\tau\mid c}
    \left(
        \mu_\tau(a_\tau,c)
    \right), \\ 
&\hspace{-7.7cm}\text{where} \\
&\hspace{-7.5cm}
    \mu_\tau(a_\tau,c)=\mathbb{E}[a\mid a_\tau,c],
    \qquad
    \Sigma_{\mathrm{div}}(a_\tau,\tau,c)
    =
    \operatorname{Cov}(a\mid a_\tau,c).
    \end{aligned}
    \label{eq:total_covariance}
\end{equation}
Equation~\eqref{eq:total_covariance} establishes a
relationship at the covariance level, not a direct decomposition of
AAC entropy. The equation shows that DUAC estimates the trajectory-conditioned covariance component that contributes to the marginal covariance underlying AAC's continuous action entropy.

The first term captures \emph{within-state uncertainty}, i.e., the ambiguity among actions remaining for a fixed $a_\tau$, while the second captures \emph{between-state uncertainty}, i.e., variation of the posterior mean across noisy trajectories. Sampled-based (distribution-level) uncertainty therefore summarizes the overall diversity of possible actions, whereas divergence-based (local) uncertainty isolates the ambiguity associated with the particular trajectory being generated.
This local characterization is particularly useful because trajectories under the same context $c$ can exhibit different levels of uncertainty. Rather than assigning a single uncertainty value to the entire conditional action distribution, $\Sigma_{\mathrm{div}}$ allows uncertainty to be evaluated along individual trajectories, revealing where the model encounters locally ambiguous decisions.

\subsection{Analysis and Ablations}

\begin{figure}[h]
  \centering
  \includegraphics[width=0.9\columnwidth]{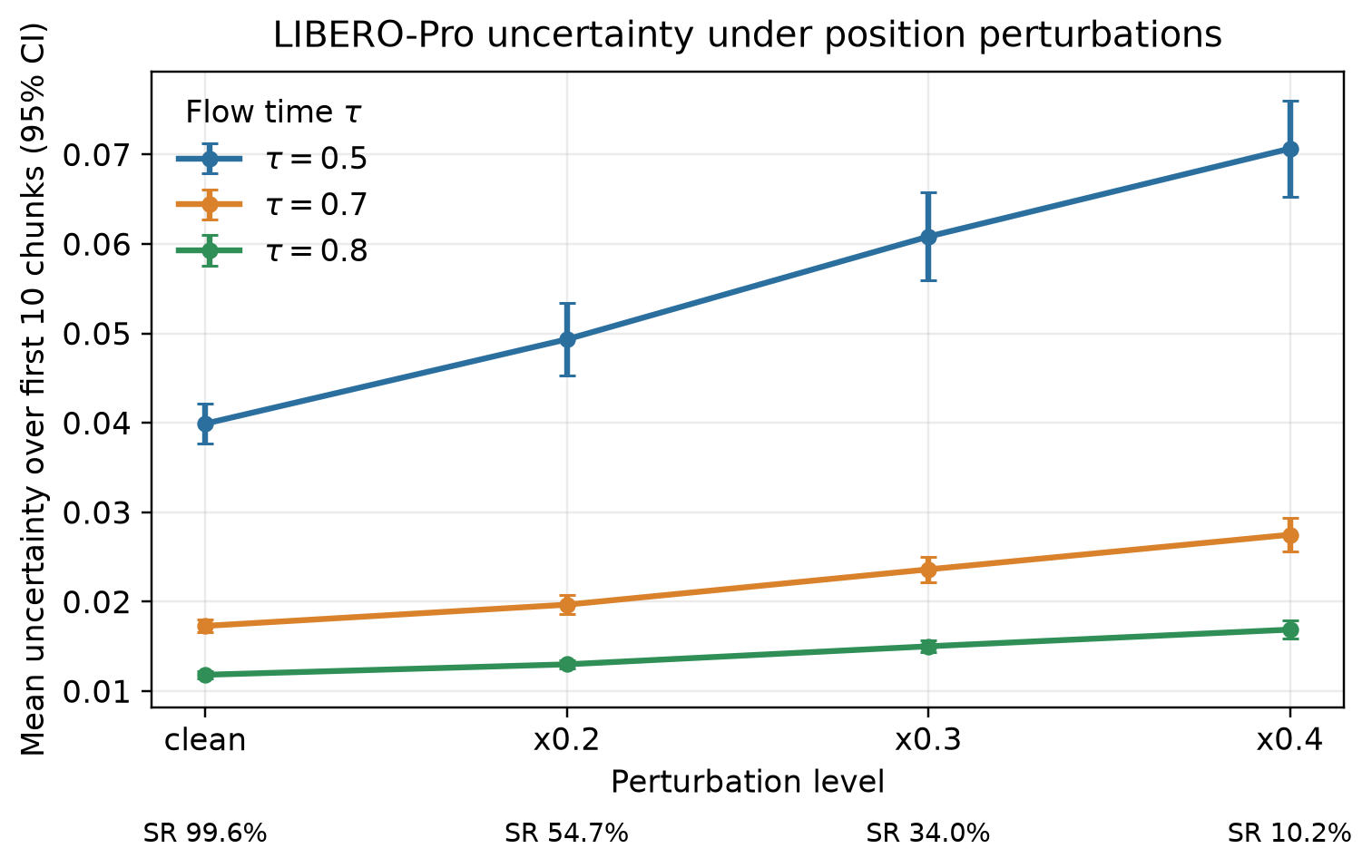}
  \caption{Divergence uncertainty under increasing LIBERO-Pro position perturbations. Curves show mean uncertainty at different flow times, error bars show 95\% confidence intervals, and success rates are shown below the perturbation levels.}
  \label{fig:libero_pro_shift_v3}
\end{figure}

\textbf{Sensitivity to Distribution-Shift Severity.}
We evaluate all 10 LIBERO-Object tasks under LIBERO-Pro position perturbations at multiple flow times.
Figure~\ref{fig:libero_pro_shift_v3} shows that uncertainty increases monotonically from the clean condition through perturbation level $x0.4$ at every evaluated flow time. Sensitivity is greater at earlier, noisier flow states, with the strongest separation at the earliest evaluated time, $\tau=0.5$. At this stage, the latent contains less information about the final action, so the vector field depends more strongly on the observation and instruction; mismatches in the conditioning are therefore more visible. Concurrently, success rate decreases from $99.6\%$ in the clean condition to $54.7\%$, $34.0\%$, and $10.2\%$ under progressively stronger perturbations. Thus, divergence uncertainty tracks the severity of this controlled position shift together with the associated degradation in policy performance.

\begin{figure}[h]
  \centering
  \includegraphics[width=\columnwidth]{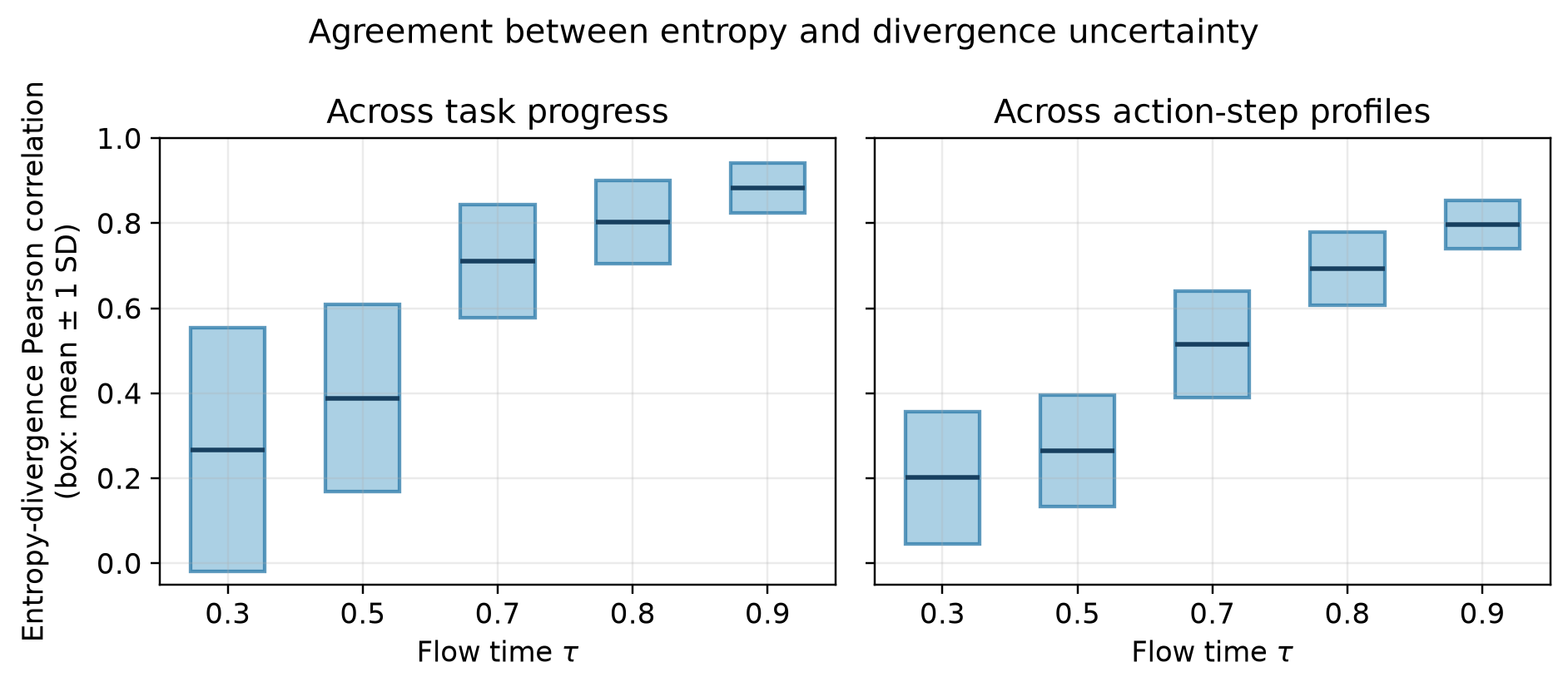}
  \caption{Pearson correlation between entropy and divergence across task progress and per-action-step uncertainty profiles. Boxes show the mean $\pm 1$ standard deviation.}
  \label{fig:entropy_divergence_v3}
\end{figure}

\textbf{Agreement with Action Entropy.}
We compare divergence uncertainty with the sampling-based action entropy used by AAC~\cite{liang2026aac}. Figure~\ref{fig:entropy_divergence_v3} reports the Pearson correlation between chunk-level scores over task progress and between per-action-step uncertainty profiles. At $\tau=0.9$, the mean correlations reach $0.88$ and $0.80$, respectively. At this later flow time, the latent is close to the generated action trajectory, and the per-step uncertainty profile is smoother and less dominated by the initial latent noise. Divergence therefore broadly recovers the uncertainty structure captured by action entropy once the trajectory has largely formed.

\textbf{Adaptive Execution under Multiple Valid Strategies.}

Table~\ref{tab:task8_decisions_v3} compares AAC and DUAC on the two-moka-pot
task, where multiple valid execution orders are available.

\begin{table}[H]
  \centering
  \footnotesize
  \setlength{\tabcolsep}{5pt}
  \begin{tabular}{lccc}
    \toprule
    Method & SR & Decisions & Mean H \\
    \midrule
    Entropy AAC & 75\% & 58.01 & 7.53 \\
    DUAC & 75\% & \textbf{49.48} & \textbf{8.83} \\
    \bottomrule
  \end{tabular}
  \caption{Execution statistics on the two-moka-pot task over 100 rollouts.}
  \label{tab:task8_decisions_v3}
\end{table}

Table~\ref{tab:robocasa_door_decisions} compares success and the mean number
of policy decisions made by AAC and DUAC on matched single- and double-door
RoboCasa tasks. DUAC improves success on all four tasks while reducing the
number of decisions by $1.48$--$5.66$ per episode. The largest differences
occur on \texttt{OpenDoubleDoor}, where DUAC improves success by $26$
percentage points and uses $5.66$ fewer decisions. This result is consistent
with DUAC avoiding some of the additional replanning induced by variation
among valid strategies.

\begin{table}[H]
  \centering
  \footnotesize
  \setlength{\tabcolsep}{3pt}
  \resizebox{\columnwidth}{!}{%
  \begin{tabular}{lccccc}
    \toprule
    Task & SR--AAC & SR--DUAC & Decisions--AAC & Decisions--DUAC & Decisions vs. AAC \\
    \midrule
    OpenSingleDoor & 50\% & 52\% & 16.13 & 14.65 & $-1.48$ \\
    OpenDoubleDoor & 36\% & 62\% & 30.54 & 24.88 & $-5.66$ \\
    CloseSingleDoor & 73\% & 83\% & 13.08 & 11.43 & $-1.65$ \\
    CloseDoubleDoor & 35\% & 37\% & 19.84 & 17.02 & $-2.82$ \\
    \bottomrule
  \end{tabular}%
  }
  \caption{Success rates and mean policy decisions per episode on matched RoboCasa door tasks. Decision differences are computed as DUAC minus AAC.}
  \label{tab:robocasa_door_decisions}
\end{table}

\subsection{Additional Control Results}

\textbf{Standard LIBERO.}
Table~\ref{tab:libero_control_v3} reports suite-level success rates on standard
LIBERO. Performance is near saturation for all methods, while \method obtains
the highest average and the strongest result on LIBERO-10.

\begin{table}[H]
  \centering
  \tiny
  \setlength{\tabcolsep}{3pt}
  \resizebox{\columnwidth}{!}{%
  \begin{tabular}{lccccc}
    \toprule
    Method & Spatial & Object & Goal & LIBERO-10 & Average \\
    \midrule
    Fixed-10 & \textbf{99.4\%} & 99.6\% & 97.8\% & 92.8\% & 97.4\% \\
    AutoHorizon & 98.1\% & 98.3\% & 97.8\% & 93.7\% & 96.98\% \\
    AAC & 98.8\% & \textbf{99.8\%} & 98.2\% & 93.0\% & 97.5\% \\
    \method (ours) & \textbf{99.4\%} & 99.2\% & \textbf{98.4\%} & \textbf{94.8\%} & \textbf{98.0\%} \\
    \bottomrule
  \end{tabular}%
  }
  \caption{Success rates (\%) with a $\pi_{0.5}$ backbone on LIBERO. 
  }
  \label{tab:libero_control_v3}
\end{table}

\textbf{LIBERO-Plus.}
Table~\ref{tab:libero_plus_control_v3} compares adaptive control across
perturbation categories. \method achieves the highest average success and
performs best in most categories, with its largest gain under camera shifts.

\begin{table}[H]
  \centering
  \footnotesize
  \setlength{\tabcolsep}{2pt}
  \begin{tabular}{lcccc}
    \toprule
    Perturbation & Fixed-10 & AutoHorizon & AAC & \method (ours) \\
    \midrule
    Background & \textbf{97.00\%} & 96.75\% & 96.45\% & 96.50\% \\
    Camera & 60.70\% & 60.75\% & 63.30\% & \textbf{63.85\%} \\
    Language & 85.20\% & 84.65\% & 84.95\% & \textbf{85.35\%} \\
    Light & 95.50\% & 95.70\% & 95.70\% & \textbf{96.15\%} \\
    Objects layout & 85.20\% & 85.15\% & 85.70\% & \textbf{85.75\%} \\
    Robot init. & \textbf{83.15\%} & 82.85\% & 80.90\% & 81.70\% \\
    Sensor noise & 85.85\% & 85.55\% & 85.15\% & \textbf{86.20\%} \\
    \midrule
    Average & 84.66\% & 84.49\% & 84.59\% & \textbf{85.07\%} \\
    \bottomrule
  \end{tabular}
  \caption{Success rates (\%) with a $\pi_{0.5}$ backbone across LIBERO-Plus perturbation categories.}
  \label{tab:libero_plus_control_v3}
\end{table}

\end{document}